\documentclass{article}

\usepackage[preprint]{neurips_2026}

\usepackage[utf8]{inputenc}
\usepackage[T1]{fontenc}
\usepackage{hyperref}
\usepackage{url}
\usepackage{booktabs}
\usepackage{amsfonts}
\usepackage{amsmath}
\usepackage{amssymb}
\usepackage{nicefrac}
\usepackage{microtype}
\usepackage{xcolor}
\usepackage{graphicx}
\usepackage{enumitem}
\usepackage{multirow}
\usepackage{tikz}
\usepackage{placeins}
\usetikzlibrary{arrows.meta,positioning,shapes.geometric}

\workshoptitle{Physical World AI: Geometry, Characteristics, and Multimodal Sensing}

\title{Token-Budgeted Candidate Retrieval for 3D Scene-Grounded VLMs:\\Category Names vs.\ Pairwise Geometry}

\author{%
Micha{\l} Zakrzewski\textsuperscript{1,3},
Mariia Shpir\textsuperscript{2,3},
Taras Rumezhak\textsuperscript{3}\\[0.35em]
\textsuperscript{1}AGH University of Krakow, Krakow, Poland\\
\textsuperscript{2}ETH Zurich\\
\textsuperscript{3}SoftServe
}

\begin{document}

\maketitle

\begin{abstract}
Under a fixed context budget, which objects and relations should a 3D scene-grounded VLM see first?
On ScanNet indoor referring benchmarks~\cite{dai2017_scannet}, matching the query to object category names outperforms pairwise geometry scoring at every reported $K$. The gaps are largest at high recall.
Analytic pairwise cues beat per-object cue scoring and query-agnostic $k$NN graphs, including on paraphrases. MiniLM and BGE-small still score higher on those splits.
We evaluate token-budgeted candidate retrieval with axis-aligned boxes, a token-ratio metric, training-free baselines, two small encoders that match $q$ to each object's category name, and QCRR (Query-Conditioned Relation Retrieval), a deterministic pairwise-geometry scorer.
Across ScanRefer gt\_val plus a custom $20\%$ scan-id split of Nr3D and Sr3D ($35{,}504$ queries; not official ReferIt3D val), additive noun matching reaches R@10 $0.864$ on ScanRefer against QCRR's $0.626$; MiniLM and BGE-small reach $0.911$ to $0.913$ on ScanRefer and $0.941$ to $0.947$ on Nr3D.
A corrected 3DGraphLLM harness (native Acc@0.25 $0.67$ on the full graph, $n{=}300$ ScanRefer queries) shows that context selection changes downstream accuracy under a matched object budget (Random $0.44$, QCRR $0.56$, MiniLM $0.67$, oracle $0.70$). Semantic retrieval beats hand-designed geometry on that harness.
We will release the protocol, retriever implementations, and per-query outputs so new selectors can be swapped in without rewriting the analysis.
\end{abstract}

\section{Introduction}

Physical-world VLMs for indoor scenes have to pick one object among dozens under a hard token budget: a user says ``the lamp on the desk.''
3DGraphLLM and related systems serialize a subset of relational triplets from a scene graph into the prompt~\cite{zemskova2025_3dgraphllm}.
The default graph is query-agnostic, so the same edges appear whether the query mentions spatial layout or only a category name.
Under that budget, category-name matching already outperforms pairwise geometry on the indoor splits we report.
Geometry still helps against Cue-only, query-agnostic $k$NN, and paraphrases.
Given $N$ axis-aligned boxes and a referring expression, which top-$K$ objects and top-$M$ relations should enter the prompt?

We contribute a protocol for measuring what drives retrieval quality:
(i)~a plug-in \texttt{retrieve($q$, boxes, names)} interface with a normalized token-ratio metric and per-query logs;
(ii)~QCRR, a deterministic geometry instrument for ablation;
(iii)~multi-axis stratifications (cue type, scene size, spatial-language flag, relation budget);
(iv)~a corrected end-to-end evaluation on 3DGraphLLM after auditing our own harness.

Category-name matching is stronger than pairwise geometry at all reported $K$ on ScanRefer gt\_val (MiniLM R@1 $0.461$ vs.\ QCRR $0.202$; noun R@10 $0.864$ vs.\ $0.626$) and on Nr3D (MiniLM R@10 $0.941$ vs.\ QCRR $0.634$).
MiniLM and BGE-small, matching $q$ to names only, reach $0.911$ and $0.913$ R@10 on ScanRefer.
Pairwise geometry helps against Cue-only and query-agnostic $k$NN.
MiniLM remains higher on ScanRefer and on the Nr3D spatial-language split.
On the $8.9\%$ paraphrased ScanRefer bucket, QCRR beats Cue-only by $+15.2$\,pp R@10 (MiniLM is not reported on that slice).
On Sr3D, QCRR R@1 is $0.254$ vs.\ $k$NN $0.016$ (below Random $0.022$); BGE leads at $0.312$, and noun matching is already near ceiling at R@10 ($0.993$).
Under a matched object budget on 3DGraphLLM, query-conditioned selection beats random and naive $k$NN, while MiniLM beats QCRR downstream, matching the retrieval ranking.

\paragraph{Scope.}
All main-table experiments use indoor ScanNet-derived benchmarks and axis-aligned boxes.
We evaluate one released 3DGraphLLM checkpoint and do not measure other backbones.

\section{Related work}

\paragraph{3D referring expression grounding.}
Trained 3D grounders fuse features and relations end-to-end~\cite{zhao2021_3dvgtransformer,huang2022_mvt}.
We isolate the upstream retrieval layer that feeds graph-serialized context to LLM-based grounders~\cite{huang2024_chatscene,zemskova2025_3dgraphllm}.

\paragraph{Training-free and LLM-driven 3D reasoning.}
LaSP~\cite{mi2025_lasp}, Transcrib3D~\cite{fang2024_transcrib3d}, and SORT3D~\cite{kang2025_sort3d} delegate spatial reasoning to programs or toolboxes at higher runtime cost.
We use a cheaper analytic scorer to measure when pairwise geometry helps over per-object heuristics.

\paragraph{Retrieval under prompt budgets.}
KeySG~\cite{mu2025_keysg} and EmbodiedRAG~\cite{xie2024_embodiedrag} retrieve subgraphs or passages under token limits.
We apply the same concern to 3D relation triplets and report a serializer-relative token-ratio metric for budget-fair comparison.
The rest of the paper asks when pairwise geometry beats category-name matching under that budget.

\section{Token-budgeted retrieval setup}
\label{sec:method}

\begin{figure}[t]
\centering
\begin{tikzpicture}[
  >=Stealth, node distance=9mm,
  every node/.style={font=\footnotesize},
  box/.style={draw, rounded corners=2pt, minimum height=7mm, minimum width=18mm,
              align=center, inner sep=2pt},
  qbox/.style={box, fill=blue!10},
  gbox/.style={box, fill=orange!15},
]
\node[qbox] (q) {Query $q$};
\node[box, right=3mm of q]  (cue)  {Cue\\parse};
\node[gbox, right=3mm of cue] (geo) {Pairwise\\geometry};
\node[box, right=3mm of geo] (score) {Cue\\scoring};
\node[box, right=3mm of score] (hub) {Hub\\down-weight};
\node[box, right=3mm of hub] (out) {Top-$M$ rels,\\top-$K$ objs};
\node[box, below=3mm of cue] (boxes) {$N$ axis-aligned boxes};

\draw[->] (q) -- (cue);
\draw[->] (cue) -- (geo);
\draw[->] (boxes.east) -- (geo.south);
\draw[->] (geo) -- (score);
\draw[->] (score) -- (hub);
\draw[->] (hub) -- (out);
\end{tikzpicture}
\caption{QCRR is a five-stage analysis instrument: parse spatial cues from $q$, score ordered object pairs with analytic geometry on axis-aligned boxes, down-weight structural hubs, then return ranked relations and candidate objects.}
\label{fig:diagram}
\end{figure}
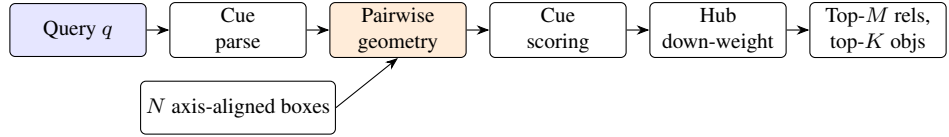

The evaluation goal is fair top-$K$ object and top-$M$ relation selection under a token ratio.
QCRR is one \texttt{retrieve()} implementation (Figure~\ref{fig:diagram}), used to ablate pairwise geometry.
MiniLM and BGE-small implement \texttt{retrieve($q$, names)} on the same boxes.

\paragraph{Inputs and outputs.}
A scene exposes $N$ axis-aligned bounding boxes (center $c_i\!\in\!\mathbb{R}^3$, size $s_i$) and category strings $n_i$.
For query $q$, a retriever returns ranked object indices and, when applicable, ranked relation pairs.
Default budgets are $M\!=\!60$ relations and $K\!=\!20$ objects, matching 3DGraphLLM evaluation.

\paragraph{QCRR scoring (analysis instrument).}
QCRR is an ablation of pairwise geometry.
Phrase-first cue matching fires multi-word triggers (\emph{in front of}, \emph{next to}) before single-word ones (\emph{left}, \emph{above}, \emph{near}).
For each ordered pair $(i,j)$ we compute offsets, distance, 3D IoU, and containment ratios.
Cue $c$ maps to an analytic term $s_c\!\in\![0,1]$ (Table~\ref{tab:cue_geom_map}); multiple cues are averaged.
Relations incident to structural categories (wall, floor, ceiling, door, window, doorframe, stair rail) are multiplied by $0.1$ before top-$M$ selection.
Object scores take the max incident pair score; a multiplicative noun prior ($\beta\!=\!1.5$) applies when the category string appears in $q$.

\begin{table}[t]
\centering
\small
\caption{Cue families mapped to geometric signals in the ScanNet world frame.
$\sigma$ is a logistic; $d_{\max}$ is the scene diagonal; $\Delta x_{ij}$ is the world-frame center offset of pair $(i,j)$.}
\label{tab:cue_geom_map}
\begin{tabular}{@{}ll@{}}
\toprule
Cue family & Signal \\
\midrule
Left / Right & $\sigma(\pm\Delta x_{ij})$ \\
Front / Behind & $\sigma(\mp\Delta y_{ij})$ \\
Above / Below & $\sigma(\pm\Delta z_{ij})$ \\
Near / Far & $1 - d_{ij}/d_{\max}$ \;/\; $d_{ij}/d_{\max}$ \\
On / Inside & vertical overlap $\times$ containment \\
Overlap & 3D IoU \\
\bottomrule
\end{tabular}
\end{table}

\paragraph{Token ratio.}
Let $\tau\!=\!L_{\text{retrieved}}/L_{\text{full}}$ where $L$ counts GPT-2~\cite{radford2019gpt2} tokens in the serialized triplet prefix.
We use $\tau$ to compare relation budgets fairly (Figure~\ref{fig:pareto}).
All reported $\tau$ use this tokenizer; we did not ablate tokenizer choice.

\section{Evaluation protocol and baselines}

\paragraph{Datasets.}
\textbf{ScanRefer}~\cite{chen2019_scanrefer}: $9{,}508$ val queries on ScanNet~\cite{dai2017_scannet}, free-form language.
ScanRefer retrieval in the main tables uses the gt\_val object-index protocol (consistent with our end-to-end harness).
\textbf{Nr3D} / \textbf{Sr3D}~\cite{achlioptas2020_referit3d}: $7{,}918$ and $18{,}078$ queries on a custom $20\%$ scan-id split (seed 42).
This is not official ReferIt3D val and not ScanNet's official val list.

\paragraph{Metrics.}
Recall@$K$ for $K\!\in\!\{1,5,10,20\}$; downstream Acc@0.25 (3D IoU between predicted and GT boxes).

\paragraph{Baselines.}
Training-free comparators in the main tables: Random, $k$NN-graph, Cue-only, Cue-only + noun (per-object cues plus an additive category bonus), and a centroid heuristic (Table~\ref{tab:scene_size}).
\textbf{Learned:} cosine similarity between $q$ and each object's category name using frozen encoders; we report MiniLM~\cite{wang2020_minilm,reimers2019sentence} (\texttt{all-MiniLM-L6-v2}, ${\sim}22$M) and BGE-small~\cite{xiao2024_cpack} (\texttt{bge-small-en-v1.5}, ${\sim}33$M) on ScanRefer, Nr3D, and Sr3D.
This is conservative: objects have no free-form caption at inference.
As an upper bound on richer object text, we mean-pool MiniLM embeddings of other ScanRefer val descriptions of the same object, excluding the current query.
All methods see the same boxes and query string.

\section{Retrieval results}
\label{sec:retrieval}

\begin{table}[t]
\centering
\small
\caption{Retrieval on ScanRefer gt\_val ($9{,}508$ queries). Bold marks best per column among listed methods.}
\label{tab:scanrefer_retrieval}
\setlength{\tabcolsep}{4pt}
\begin{tabular}{@{}lcccc@{}}
\toprule
Method & R@1 & R@5 & R@10 & R@20 \\
\midrule
Random & 0.032 & 0.163 & 0.322 & 0.606 \\
$k$NN-graph & 0.027 & 0.170 & 0.326 & 0.609 \\
Cue-only + noun & 0.260 & 0.673 & 0.818 & 0.924 \\
QCRR & 0.202 & 0.493 & 0.626 & 0.844 \\
Learned (MiniLM) & \textbf{0.461} & 0.826 & 0.911 & 0.973 \\
Learned (BGE-small) & 0.455 & \textbf{0.830} & \textbf{0.913} & \textbf{0.974} \\
Noun (additive, gt\_val) & 0.255 & 0.728 & 0.864 & 0.946 \\
\bottomrule
\end{tabular}
\end{table}

\begin{table}[t]
\centering
\small
\caption{Cross-benchmark retrieval. Nr3D/Sr3D use a custom $20\%$ scan-id split (seed 42), not official ReferIt3D val; ScanRefer is gt\_val, matching Table~\ref{tab:scanrefer_retrieval}. MiniLM and BGE-small match $q$ to category names only.}
\label{tab:cross_benchmark}
\setlength{\tabcolsep}{3.0pt}
\begin{tabular}{@{}lcccc|cccc|cccc@{}}
\toprule
& \multicolumn{4}{c|}{ScanRefer} & \multicolumn{4}{c|}{Nr3D} & \multicolumn{4}{c}{Sr3D} \\
Method & R@1 & R@5 & R@10 & R@20 & R@1 & R@5 & R@10 & R@20 & R@1 & R@5 & R@10 & R@20 \\
\midrule
Random & .032 & .163 & .322 & .606 & .028 & .161 & .321 & .597 & .022 & .119 & .226 & .462 \\
$k$NN-graph & .027 & .170 & .326 & .609 & .031 & .180 & .332 & .603 & .016 & .094 & .200 & .451 \\
Cue-only + noun & .260 & .673 & .818 & .924 & .212 & .735 & .856 & .935 & .267 & .960 & .993 & .999 \\
QCRR & .202 & .493 & .626 & .844 & .182 & .524 & .634 & .820 & .254 & .599 & .616 & .754 \\
MiniLM & .461 & .826 & .911 & .973 & .271 & .857 & .941 & .973 & .262 & .976 & .996 & .999 \\
BGE-small & .455 & .830 & .913 & .974 & .273 & .863 & .947 & .978 & .312 & .979 & .998 & 1.00 \\
\bottomrule
\end{tabular}
\end{table}

\begin{figure}[t]
\centering
\includegraphics[width=0.52\linewidth]{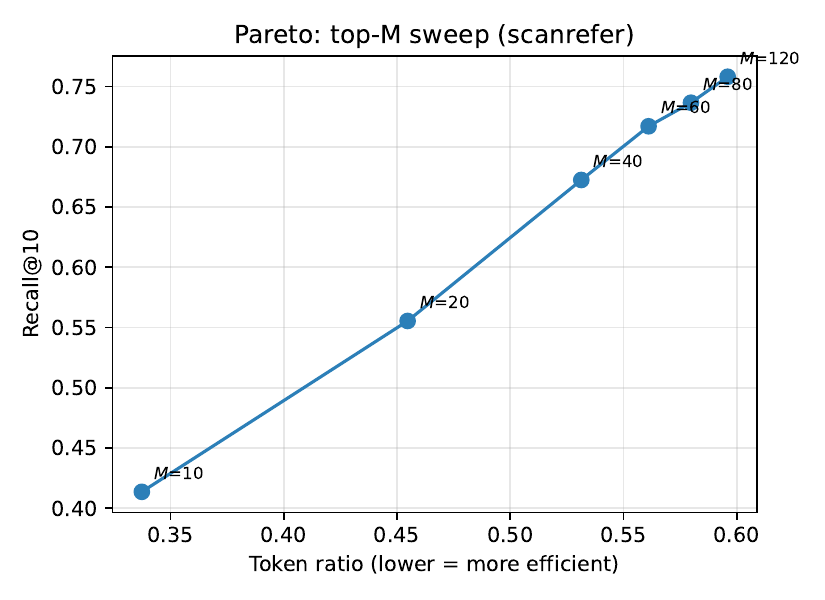}
\caption{ScanRefer Recall@10 vs.\ token ratio as relation budget $M$ varies. Main tables use $M{=}60$.}
\label{fig:pareto}
\end{figure}

Table~\ref{tab:scanrefer_retrieval} summarizes ScanRefer gt\_val.
MiniLM R@1 is $0.461$ vs.\ QCRR $0.202$; both category-name encoders and the additive noun baseline exceed QCRR by $\approx\!+29$ and $+24$\,pp at R@10.
MiniLM and BGE-small differ by at most $0.6$\,pp on this split (the Sr3D R@1 gap is $5.0$\,pp).
Removing QCRR's noun prior drops R@10 by $19.8$\,pp (Table~\ref{tab:ablation}), so much of QCRR's own recall is lexical.
The same ranking holds on Nr3D (Table~\ref{tab:cross_benchmark}): MiniLM R@10 $0.941$ vs.\ QCRR $0.634$.
Giving MiniLM leave-one-out descriptions of the same instance (Table~\ref{tab:learned_text}) lifts R@1 from $0.461$ to $0.587$ and R@10 from $0.911$ to $0.971$.
Leave-one-out captions are unavailable at deployment. They show how far semantic retrieval can go with richer object text.
On Sr3D, noun matching is already near ceiling at R@10 ($0.993$); BGE still leads at R@1 ($0.312$ vs.\ Cue-only + noun $0.267$ and QCRR $0.254$).
QCRR R@1 is $15.9{\times}$ above query-agnostic $k$NN-graph ($0.254$ vs.\ $0.016$), which is below Random ($0.022$). That comparison isolates pairwise geometry from a query-blind graph. MiniLM and BGE remain higher on Sr3D at every $K$.

\begin{table}[t]
\centering
\small
\caption{MiniLM on ScanRefer gt\_val. Leave-one-out uses other val descriptions of the same object, not the current query.}
\label{tab:learned_text}
\begin{tabular}{@{}lcccc@{}}
\toprule
Object text & R@1 & R@5 & R@10 & R@20 \\
\midrule
Category name & 0.461 & 0.826 & 0.911 & 0.973 \\
Leave-one-out descriptions & \textbf{0.587} & \textbf{0.914} & \textbf{0.971} & \textbf{0.996} \\
\bottomrule
\end{tabular}
\end{table}

\paragraph{When geometry matters.}
On ScanRefer, $91.1\%$ of queries name the target category verbatim; there Cue-only leads at R@10 ($0.756$ vs.\ QCRR $0.652$ on that subset).
On the $8.9\%$ paraphrased-target bucket, QCRR R@10 is $0.440$ vs.\ Cue-only $0.288$ ($+15.2$\,pp); Table~\ref{tab:synonym} shows the ranking flips at every $K$ once verbatim matching fails.
MiniLM and BGE-small are not in that table, so this is a Cue-only comparison.
On Nr3D we split by the ReferIt3D CSV flag \texttt{uses\_spatial\_lang} ($7{,}140$ vs.\ $778$; Table~\ref{tab:view_split}), which is not the official view-dependence split.
QCRR R@1 is higher on the no-spatial-language slice ($0.244$ vs.\ $0.177$).
Cue-only is flat; MiniLM rises more ($0.260$ vs.\ $0.369$) and wins every $K$ on both slices.

\begin{table}[t]
\centering
\small
\caption{ScanRefer gt\_val stratified by verbatim category match. Non-exact queries use synonyms, hypernyms, or paraphrases (e.g., ``wooden cabinets'' for \texttt{kitchen cabinets}).}
\label{tab:synonym}
\setlength{\tabcolsep}{3pt}
\begin{tabular}{@{}lrcccccc@{}}
\toprule
& & \multicolumn{3}{c}{QCRR} & \multicolumn{3}{c}{Cue-only} \\
Bucket & \#queries & R@1 & R@5 & R@10 & R@1 & R@5 & R@10 \\
\midrule
Exact-match & 8{,}660 & 0.251 & 0.529 & 0.652 & 0.231 & \textbf{0.592} & \textbf{0.756} \\
Non-exact   &   848 & \textbf{0.037} & \textbf{0.236} & \textbf{0.440} & 0.018 & 0.103 & 0.288 \\
\bottomrule
\end{tabular}
\end{table}

A second axis is instance crowding under Mask3D~\cite{schult2023_mask3d} Acc@0.25, distinct from the retrieval exact/non-exact split (Table~\ref{tab:per_class_compact}).
Chair queries alone are $23\%$ of ScanRefer val with disambiguation gap $\delta\!=\!1{-}S2/S4\!=\!0.79$ (lexical Acc@0.25 $0.15$ vs.\ class oracle $0.74$ on Mask3D boxes); bathtub and counter queries sit at $\delta\!\le\!0.09$ where category matching already nears the per-class ceiling.
The full NYU40~\cite{silberman2012indoor} breakdown is in Appendix~\ref{sec:appendix_per_class}.

\begin{table}[t]
\centering
\small
\caption{Per-class disambiguation on ScanRefer val (Mask3D Acc@0.25, not retrieval R@$K$). $S2$: lexical; $S4$: class oracle; $\delta\!=\!1{-}S2/S4$.}
\label{tab:per_class_compact}
\setlength{\tabcolsep}{4pt}
\begin{tabular}{@{}lrrrr@{}}
\toprule
Class & \#queries & $S2$ & $S4$ & $\delta$ \\
\midrule
chair & 2{,}196 & 0.15 & 0.74 & \textbf{0.79} \\
cabinet & 781 & 0.23 & 0.56 & 0.60 \\
table & 768 & 0.32 & 0.58 & 0.45 \\
counter & 84 & 0.71 & 0.79 & 0.09 \\
bathtub & 50 & 0.62 & 0.62 & 0.00 \\
\bottomrule
\end{tabular}
\end{table}

\paragraph{Scene size and budget.}
Recall@10 falls with object count $N$ for every tabulated method (Table~\ref{tab:scene_size}); MiniLM and BGE-small are not in this table.
At $N\!\ge\!61$, Cue-only + noun retains R@10 $0.609$ vs.\ QCRR $0.487$, so category matching still leads in large scenes.
Sweeping relation budget $M$ on ScanRefer shows a knee near $M\!=\!60$ (Figure~\ref{fig:pareto}); an adaptive policy $M\!=\!0.05N^2$ Pareto-dominates fixed $M\!=\!60$ on the leftover ScanRefer val protocol in Appendix~\ref{sec:appendix_adaptive_m}, not on gt\_val.

\begin{table}[t]
\centering
\small
\caption{Recall@10 by scene size on ScanRefer gt\_val.}
\label{tab:scene_size}
\begin{tabular}{@{}lrrccc@{}}
\toprule
$N$ bin & \#queries & $\bar{N}$ & QCRR & Centroid & Cue-only + noun \\
\midrule
1--20 & 1{,}061 & 16 & 0.926 & 0.685 & 0.951 \\
21--40 & 4{,}932 & 30 & 0.781 & 0.391 & 0.864 \\
41--60 & 2{,}549 & 48 & 0.594 & 0.217 & 0.741 \\
61+ & 966 & 78 & 0.487 & 0.174 & 0.609 \\
\bottomrule
\end{tabular}
\end{table}

\begin{table}[t]
\centering
\small
\caption{Nr3D split by ReferIt3D \texttt{uses\_spatial\_lang} (not official view-dependence). MiniLM matches $q$ to names.}
\label{tab:view_split}
\setlength{\tabcolsep}{4pt}
\begin{tabular}{@{}lcccc|cccc@{}}
\toprule
& \multicolumn{4}{c|}{Spatial lang.\ ($7{,}140$)} & \multicolumn{4}{c}{No spatial lang.\ ($778$)} \\
Method & R@1 & R@5 & R@10 & R@20 & R@1 & R@5 & R@10 & R@20 \\
\midrule
$k$NN-graph & 0.031 & 0.181 & 0.335 & 0.602 & 0.036 & 0.170 & 0.306 & 0.609 \\
Cue-only & 0.163 & 0.554 & 0.717 & 0.864 & 0.162 & 0.526 & 0.716 & 0.870 \\
QCRR & 0.177 & 0.520 & 0.627 & 0.829 & 0.244 & 0.558 & 0.647 & 0.748 \\
Learned (MiniLM) & \textbf{0.260} & \textbf{0.850} & \textbf{0.939} & \textbf{0.973} & \textbf{0.369} & \textbf{0.918} & \textbf{0.961} & \textbf{0.976} \\
\bottomrule
\end{tabular}
\end{table}

\begin{table}[t]
\centering
\small
\caption{Dominant cue type on ScanRefer gt\_val (QCRR Recall@10). Vertical cues are easiest in the gravity-aligned ScanNet frame.}
\label{tab:cue_breakdown}
\begin{tabular}{@{}lrr@{}}
\toprule
Cue bucket & \#queries & R@10 \\
\midrule
left/right & 3{,}957 & 0.696 \\
near/far & 3{,}155 & 0.733 \\
vertical (above/below) & 1{,}303 & 0.773 \\
non-directional (near/on/overlap) & 1{,}091 & 0.681 \\
\bottomrule
\end{tabular}
\end{table}

Horizontal cues trail vertical ones (Table~\ref{tab:cue_breakdown}). ScanNet's world $x$-axis is not aligned with the room; PCA on a leftover ScanRefer val left/right slice ($n{=}4{,}272$) gains $+2$ to $5$\,pp.
That slice is not the gt\_val cue table ($n{=}3{,}957$; Appendix~\ref{sec:appendix_canon}).
Two queries in Table~\ref{tab:cue_breakdown} have no assigned bucket ($9{,}506$ vs.\ $9{,}508$).

\paragraph{Hybrid.}
On leftover ScanRefer val (object-description matching; not gt\_val), noun-restricted pairwise scoring lifts R@10 from $0.631$ to $0.828$ and lowers $\tau$ (Appendix~\ref{sec:appendix_hybrid}).
Hybrid trails MiniLM ($0.911$) and additive noun matching ($0.864$) on gt\_val.

\section{End-to-end evaluation on 3DGraphLLM}
\label{sec:e2e}

Retrieval recall does not by itself predict downstream grounding accuracy.
We wire each retriever into the native 3DGraphLLM graph pathway (subset \texttt{assigned\_ids} under a matched object budget).
Relations enter the embedding pathway as graph edges.
We score predictions in the model's object-index space and load per-object features with the correct scene-key format.

\paragraph{Harness audit.}
Our initial submission reported Random, $k$NN, and QCRR as statistically indistinguishable downstream.
Re-auditing the harness found three bugs: relations were passed as prose instead of through the embedding pathway; labels used raw ScanNet ids instead of model object indices; object features were zero-filled from a key-format mismatch.
The released checkpoint loads completely (448 LoRA adapters, zero unexpected keys; the previously reported ``missing'' keys are frozen base weights loaded from the pretrained backbone).
After correction, native full-graph evaluation reaches Acc@0.25 $0.67$ on $n{=}300$ ScanRefer queries (seed 42), not on the full $9{,}508$-query val set.

\begin{table}[!b]
\centering
\small
\caption{Downstream Acc@0.25 on ScanRefer gt\_val ($n{=}300$, seed 42) under a matched object budget. GT-in-budget is a separate column. Full graph included as an unconstrained reference.}
\label{tab:e2e}
\setlength{\tabcolsep}{4pt}
\begin{tabular}{@{}lcc@{}}
\toprule
Context selector & Acc@0.25 & GT in budget \\
\midrule
Oracle (GT-incident relations first) & 0.697 & 1.00 \\
Learned retriever (MiniLM) & 0.673 & 0.96 \\
Full graph (no budget) & 0.670 & 1.00 \\
Noun baseline & 0.647 & 0.94 \\
QCRR & 0.557 & 0.82 \\
Random & 0.443 & 0.60 \\
$k$NN-graph selection & 0.410 & 0.57 \\
\bottomrule
\end{tabular}
\end{table}

Table~\ref{tab:e2e} shows that Acc@0.25 spans $0.41$ to $0.70$ under the same backbone and budget convention.
Query conditioning helps: QCRR beats Random by $+0.114$ (95\% CI $[0.034, 0.193]$ on $n{=}300$; pooled over two seeds at $n{=}600$: $+0.110$ $[0.054, 0.166]$) and beats $k$NN-graph selection by $+0.147$ $[0.068, 0.226]$.
Semantic selectors beat geometry: MiniLM and noun baselines exceed QCRR by $+0.116$ $[0.039, 0.194]$ and $+0.090$ $[0.012, 0.168]$ respectively, matching the retrieval ranking.
Oracle ($0.697$) is not statistically above full graph ($0.670$) at this sample size. Remaining errors may come from more than the object budget.

\paragraph{Mechanism.}
GT-in-budget rate explains much of the spread (Random $0.60$, QCRR $0.82$, learned $0.96$, oracle $1.00$).
Improving retrieval without placing the target in the serialized graph has limited downstream effect.

\section{Component ablation}
\label{sec:ablation}

\begin{table}[!b]
\centering
\small
\caption{QCRR ablations on ScanRefer gt\_val ($9{,}508$ queries), same protocol as Table~\ref{tab:scanrefer_retrieval}.}
\label{tab:ablation}
\begin{tabular}{@{}lcccc@{}}
\toprule
Config & R@1 & R@5 & R@10 & R@20 \\
\midrule
A0: Full QCRR (max-edge) & 0.202 & 0.493 & 0.626 & 0.844 \\
A2: $-$ hub penalty & 0.172 & 0.446 & 0.588 & 0.825 \\
A3: $-$ noun boost & 0.048 & 0.231 & 0.428 & 0.675 \\
A6: sum\_top3 aggregator & 0.169 & 0.377 & 0.536 & 0.835 \\
\bottomrule
\end{tabular}
\end{table}

Table~\ref{tab:ablation} uses the same gt\_val object-index protocol as the main retrieval table; config ids follow the code release (A1, A4, A5 omitted).
Removing the noun boost costs $19.8$\,pp at R@10; hub suppression is worth $3.8$\,pp.
Replacing max-edge aggregation with a sum of the top three incident relations (A6) hurts R@10 ($-9.0$\,pp).

\section{Discussion and limitations}

\paragraph{Takeaways for physical-world VLMs.}
On indoor queries that name a category, object labels and semantic matching move R@$K$ more than pairwise geometry.
Analytic geometry still beats Cue-only and query-agnostic $k$NN, including on paraphrases. MiniLM is higher at R@1 on ScanRefer and on the Nr3D spatial-language split.
Relation budget can track scene size ($M\!\propto\!N^2$). Horizontal cues may need a room axis (Appendix~\ref{sec:appendix_canon}).

\paragraph{A when-to-use-geometry rule.}
Use this checklist before adding pairwise geometry to a context builder:
(i)~If the GT category string is absent from the query (Table~\ref{tab:synonym}), geometry or a hybrid beats per-object Cue-only scoring. MiniLM is not reported on that slice.
(ii)~If the target class has a wide same-instance pool ($\delta\!\gtrsim\!0.4$ in Table~\ref{tab:per_class_compact}), narrow by category first. Hybrid beats full pairwise QCRR on leftover ScanRefer val (Table~\ref{tab:hybrid}) and trails MiniLM on gt\_val.
(iii)~On the Nr3D no-spatial-language slice, world-frame pairwise scoring beats Cue-only at R@1 (Table~\ref{tab:view_split}); MiniLM is still higher on that split.
(iv)~If the class is essentially unique ($\delta\!\lesssim\!0.1$), stop at semantics.

\paragraph{Failure modes.}
\textbf{Paraphrase + multi-instance.}
A ScanRefer query asks for ``brown wooden cabinets on the side of the kitchen'' while the GT label is \texttt{kitchen cabinets}; several \texttt{cabinet} detections co-occur.
QCRR's noun prior lifts all cabinet-tagged boxes equally, so pairwise ``beside'' scores among dining chairs can outrank the correct kitchen run unless the shortlist is noun-restricted first.
\textbf{Uncanonicalized horizontal cues.}
On gt\_val left/right queries, R@10 ($0.696$) trails vertical cues ($0.773$) because ScanNet's world $x$-axis is not aligned with the room's dominant wall.
PCA axis alignment recovers $+2.7$ to $4.7$\,pp on a different ScanRefer val left/right slice ($n{=}4{,}272$; Table~\ref{tab:canon}). We leave that $\Delta$ off Table~\ref{tab:cue_breakdown}.

\paragraph{Multi3DRef.}
On Multi3DRef val (Table~\ref{tab:multi3dref}), lexical class-matching F1@0.25 is $0.322$ (ST-wo-d), $0.185$ (ST-w-d), $0.695$ (ZT-wo-d), $0.421$ (ZT-w-d); pooled $0.281$. Random stays near zero.

\paragraph{Limitations.}
Axis-aligned boxes misrepresent elongated or articulated objects; view-dependent language is only crudely handled in a fixed world frame.
We compare two small category-name encoders (ScanRefer, Nr3D, and Sr3D) and one 3DGraphLLM checkpoint.
On CityRefer val~\cite{miyanishi2023_cityrefer} ($n{=}5{,}934$ queries, six scenes, frozen indoor cues and hubs, no landmarks) MiniLM R@1 is $0.013$, QCRR $0.008$, noun $0.011$, Random $0.004$: near chance, so we omit outdoor numbers from the main tables.
Additional scene-LLM backbones are untested.
The leave-one-out caption bound uses other ScanRefer annotations of the same instance and is not available at deployment.
The token-ratio metric depends on serialization template; absolute values are not comparable across templates.

\paragraph{Reproducibility.}
We will release retriever code, the plug-in \texttt{retrieve()} interface, per-query logs (ranks, token counts, selected relations), corrected end-to-end scripts, and a checkpoint verification utility.

\FloatBarrier
\bibliographystyle{plainnat}
\bibliography{references}

\appendix

\section{Implementation details}
\label{sec:appendix_impl}

$M\!=\!60$, $K\!=\!20$, noun boost $\beta\!=\!1.5$, hub penalty $\times 0.1$ on wall/floor/ceiling/door/window/doorframe/stair rail.
Cue vocabulary and phrase-first priority are listed in our code release.
QCRR is $O(N^2)$ on the pair loop; logged wall time is $49$\,ms/query on Nr3D and $64$\,ms/query on Sr3D on a single CPU core.
Category-name MiniLM/BGE on the Nr3D/Sr3D split in Table~\ref{tab:cross_benchmark} is \texttt{scripts/eval\_referit3d\_learned.py}.

\section{Noun-restricted hybrid}
\label{sec:appendix_hybrid}

Leftover ScanRefer val (object-description protocol; not gt\_val in Table~\ref{tab:scanrefer_retrieval}):

\begin{table}[h]
\centering
\small
\caption{Noun-restricted hybrid vs.\ full pairwise QCRR on leftover ScanRefer val.}
\label{tab:hybrid}
\begin{tabular}{@{}lccccc@{}}
\toprule
Method & R@1 & R@5 & R@10 & R@20 & Token ratio \\
\midrule
QCRR (full pairwise) & 0.230 & 0.501 & 0.631 & 0.848 & 0.081 \\
Noun-restricted hybrid & \textbf{0.297} & \textbf{0.718} & \textbf{0.828} & \textbf{0.911} & \textbf{0.071} \\
\bottomrule
\end{tabular}
\end{table}

\section{Adaptive relation budget}
\label{sec:appendix_adaptive_m}

$M\!=\!0.05N^2$ improves leftover ScanRefer val R@10 to $0.662$ at token ratio $0.052$, vs.\ fixed $M\!=\!60$ at R@10 $0.631$ and ratio $0.081$ (same protocol as Table~\ref{tab:hybrid}, not gt\_val).

\section{End-to-end run notes}
\label{sec:appendix_e2e}

Corrected evaluation uses \texttt{qcrr/eval\_e2e\_fixed.py} with modes \texttt{full}, \texttt{budget}, \texttt{compare}, and \texttt{compare\_learned}.
Table~\ref{tab:e2e} reports seed 42 ($n{=}300$); seed 7 at $n{=}300$ gives QCRR $0.51$, Random $0.40$, oracle $0.67$, full $0.63$, $k$NN-graph $0.44$ (Random vs.\ $k$NN-graph reverses relative to seed 42).
Pooled $n{=}600$ (seeds 42 and 7): full $0.648$, QCRR $0.533$, Random $0.423$, oracle $0.682$.

\section{Synonym stratification}
\label{sec:appendix_synonym}

Table~\ref{tab:synonym} in the main text reports the full split.
Sr3D and Nr3D typically \emph{name} the target class and disambiguate among same-class instances (exact category string, multi-instance crowding).
That is a different regime from Table~\ref{tab:synonym}'s non-exact bucket (GT category string absent).
Query-agnostic $k$NN-graph fails at Sr3D R@1. Category matching still leads (BGE $0.312$ vs.\ QCRR $0.254$).

\section{Room-axis canonicalization}
\label{sec:appendix_canon}

ScanNet room orientation is not canonicalized, so ``left'' and ``right'' depend on an arbitrary world frame.
We apply a training-free fix: 2D PCA on object centers in the $x$-$y$ plane, rotate so the principal axis aligns with world $x$, then re-score left/right cues.

\begin{table}[h]
\centering
\small
\caption{PCA room-axis alignment on ScanRefer val, left/right queries only ($n{=}4{,}272$).}
\label{tab:canon}
\begin{tabular}{@{}lccc@{}}
\toprule
Metric & Original & Canonicalized & $\Delta$ \\
\midrule
R@1  & 0.178 & 0.205 & $+0.027$ \\
R@5  & 0.424 & 0.471 & $+0.047$ \\
R@10 & 0.588 & 0.616 & $+0.027$ \\
\bottomrule
\end{tabular}
\end{table}

The gain is on this val slice. It does not close the gt\_val horizontal/vertical gap in Table~\ref{tab:cue_breakdown}.
We keep canonicalization out of the default pipeline: clutter-heavy scenes can mis-estimate the principal axis. A layout estimator would be a better source of room axes.

\section{Lexical matching on Multi3DRef}
\label{sec:appendix_multi3dref}

The category-matching dominance in Section~\ref{sec:retrieval} is established on three single-target ScanNet referring benchmarks (ScanRefer, Nr3D, Sr3D).
Multi3DRef~\cite{zhang2023_multi3drefer} adds zero-target (\emph{ZT}) and multi-target (\emph{MT}) queries.
We re-run lexical baselines on Multi3DRef val under F1@0.25 with Mask3D-detected boxes (source: \texttt{qcrr-submission/results/multi3dref\_leakage.json}).

\begin{table}[h]
\centering
\small
\caption{Multi3DRef val ($11{,}120$ queries). S0: random; S2: lexical class match (grounding-level).}
\label{tab:multi3dref}
\setlength{\tabcolsep}{6pt}
\begin{tabular}{@{}lrcccc@{}}
\toprule
Subset & \#queries & S0 F1@0.25 & S2 F1@0.25 & S0 F1@0.50 & S2 F1@0.50 \\
\midrule
ST-wo-d  & 2{,}099 & 0.022 & 0.322 & 0.020 & 0.289 \\
ST-w-d   & 5{,}358 & 0.025 & 0.185 & 0.021 & 0.167 \\
MT       & 2{,}757 & 0.037 & 0.312 & 0.032 & 0.286 \\
ZT-w-d   &    378 & 0.000 & 0.421 & 0.000 & 0.421 \\
ZT-wo-d  &    528 & 0.000 & \textbf{0.695} & 0.000 & \textbf{0.695} \\
\midrule
All      & 11{,}120 & 0.026 & 0.281 & 0.022 & 0.255 \\
\bottomrule
\end{tabular}
\end{table}

Table~\ref{tab:multi3dref} reports all five subsets.
On zero-target queries where the referenced class is absent, the lexical heuristic correctly returns empty on $69.5\%$ of ZT-wo-d queries (F1@0.25 $0.695$ vs.\ $0$ for query-agnostic random).
Any retriever that cannot abstain saturates at zero on this subset regardless of spatial reasoning.
We did not run QCRR under the F1 protocol because it does not yet emit empty answers when no class matches.

\section{Per-class disambiguation difficulty}
\label{sec:appendix_per_class}

\begin{table}[h]
\centering
\small
\caption{Per-class breakdown of category-matching difficulty on ScanRefer val ($9{,}508$ queries, Mask3D-detected boxes). $\%$Unique is the fraction of queries on which the target class has a single instance. $S2$ is lexical Acc@0.25; $S4$ is the class-oracle ceiling; $\delta\!=\!1{-}S2/S4$.}
\label{tab:per_class}
\setlength{\tabcolsep}{6pt}
\begin{tabular}{@{}lrrccc@{}}
\toprule
Class & \#queries & $\%$Unique & $S2$ & $S4$ & $\delta$ \\
\midrule
chair            & 2{,}196 &   3.9 & 0.153 & 0.736 & \textbf{0.79} \\
cabinet          &    781  &  17.3 & 0.225 & 0.565 & 0.60 \\
table            &    768  &  31.0 & 0.319 & 0.581 & 0.45 \\
door             &    520  &  49.6 & 0.271 & 0.396 & 0.32 \\
window           &    366  &  41.3 & 0.372 & 0.604 & 0.38 \\
desk             &    238  &  39.5 & 0.244 & 0.433 & 0.44 \\
sink             &    177  &  67.8 & 0.186 & 0.203 & 0.08 \\
picture          &    175  &  52.6 & 0.394 & 0.623 & 0.37 \\
sofa             &    169  &  39.1 & 0.349 & 0.485 & 0.28 \\
bed              &    136  &  64.0 & 0.272 & 0.346 & 0.21 \\
bookshelf        &    125  &  47.2 & 0.328 & 0.504 & 0.35 \\
toilet           &    117  &  76.1 & 0.282 & 0.487 & 0.42 \\
curtain          &     94  &  57.4 & 0.447 & 0.500 & 0.11 \\
counter          &     84  &  88.1 & 0.714 & 0.786 & 0.09 \\
refrigerator     &     76  &   0.0 & 0.408 & 0.513 & 0.20 \\
shower curtain   &     58  & 100.0 & 0.897 & 0.983 & 0.09 \\
bathtub          &     50  & 100.0 & 0.620 & 0.620 & \textbf{0.00} \\
others (NYU40 outside the 17 above) & 3{,}378 &  1.4 & 0.286 & 0.504 & 0.43 \\
\midrule
All                                  & 9{,}508 & 18.5 & 0.268 & 0.555 & 0.52 \\
\bottomrule
\end{tabular}
\end{table}

Table~\ref{tab:per_class} decomposes ScanRefer val by GT object class (source: \texttt{qcrr-submission/results/per\_class\_leakage.json}, probe v3).
Bathroom fixtures and counters ($\delta\!\le\!0.11$) are essentially solved by category matching alone.
Chair, cabinet, table, and the long-tail \emph{others} bucket ($\delta\!\ge\!0.43$) cover $74.9\%$ of queries and set the aggregate $\delta$ to $0.52$.
Under the corrected end-to-end harness (Table~\ref{tab:e2e}), both QCRR and $k$NN-graph sit far below the per-class oracle in aggregate; semantic retrievers close more of the gap because they place the GT in budget more often.

\end{document}